\documentclass[conference]{IEEEtran}

\IEEEoverridecommandlockouts
\usepackage{cite}

\usepackage{tikz}
\usepackage{pgfplots}
\pgfplotsset{compat=1.17}
\usepackage{multirow}
\usepackage{float}
\usepackage{multicol}
\usepackage{caption}
\usepackage{tabularx}
\usepackage{amsmath} 
\usepackage{booktabs}
\usepackage{amsmath,amssymb,amsfonts}
\usepackage{algorithmic}
\usepackage{graphicx}
\usepackage{textcomp}
\usepackage{xcolor}
\usepackage{graphicx}
\usepackage{lipsum}
\usepackage{dblfloatfix} 
 \usepackage{multirow}
\usepackage{graphicx}
\usepackage{adjustbox}
\usepackage{tabularx}
\usepackage{cite}
\usepackage{amsmath,amssymb,amsfonts}
\usepackage{algorithmic}
\usepackage{graphicx}
\usepackage{textcomp}
\usepackage{xcolor}
\usepackage{booktabs}
\usepackage{graphicx}   
\usepackage{subcaption}  
\usepackage{float}       
\usepackage{array}
\def\BibTeX{{\rm B\kern-.05em{\sc i\kern-.025em b}\kern-.08em
    T\kern-.1667em\lower.7ex\hbox{E}\kern-.125emX}}
\usepackage{listings}
\usepackage{xcolor}
\usepackage{booktabs}
\usepackage{pifont}
\usepackage{makecell} 

\usepackage{cite}
\usepackage{amsmath,amssymb,amsfonts}
\usepackage{algorithm}
\usepackage{algorithmic}
\usepackage{graphicx}
\usepackage{textcomp}
\usepackage{xcolor}
\usepackage{booktabs}
\usepackage{multirow}
\usepackage{url}
\usepackage{listings}
\usepackage{array}
\usepackage{adjustbox} 

\def\BibTeX{{\rm B\kern-.05em{\sc i\kern-.025em b}\kern-.08em
    T\kern-.1667em\lower.7ex\hbox{E}\kern-.125emX}}

\begin{document}

\title{Chain of Simulation: A Dual-Mode Reasoning Framework for Large Language Models with Dynamic Problem Routing
}

\author{\IEEEauthorblockN{Saeid Sheikhi}
\IEEEauthorblockA{
Faculty of Information Technology and Electrical Engineering\\
University of Oulu, Oulu, Finland, 90014\\
Email: firstname.lastname@oulu.fi}}

\maketitle

\begin{abstract}
We present Chain of Simulation (CoS), a novel dual-mode reasoning framework that dynamically routes problems to specialized reasoning strategies in Large Language Models (LLMs). Unlike existing uniform prompting approaches, CoS employs three distinct reasoning modes: (1) computational flow with self-consistency for mathematical problems, (2) symbolic state tracking with JSON representations for spatial reasoning, and (3) hybrid fact-extraction for multi-hop inference. Through comprehensive evaluation on GSM8K, StrategyQA, and bAbI benchmarks using four state-of-the-art models (Gemma-3 27B, LLaMA-3.1 8B, Mistral 7B, and Qwen-2.5 14B), we demonstrate that CoS achieves 71.5\% accuracy on GSM8K (1.0\% absolute improvement), 90.0\% on StrategyQA (2.5\% improvement), and 19.0\% on bAbI (65.2\% relative improvement) compared to the strongest baselines. The analysis reveals that problem-specific mode selection is crucial, with computational mode achieving 81.2\% accuracy when correctly applied to mathematical problems, while misrouting leads to 0\% accuracy. We provide detailed algorithms for mode selection, state tracking, and answer extraction, establishing CoS as an effective approach for improving LLM reasoning without additional training. The framework provides superior trade-offs between accuracy and efficiency compared to Self-Consistency, achieving comparable performance at 54\% lower computational cost.
\end{abstract}


\section{Introduction}

Large Language Models (LLMs) have demonstrated remarkable capabilities in various reasoning tasks, fundamentally transforming how we approach complex problem-solving in artificial intelligence \cite{wei2022chain,wang2023selfconsistency,kojima2022zero}. However, their performance remains inconsistent across different problem types, with significant variations in accuracy depending on the nature of the reasoning required. Current prompting strategies, including the widely-adopted Chain-of-Thought (CoT) \cite{wei2022chain} and its numerous variants, apply uniform reasoning patterns regardless of problem characteristics, potentially limiting their effectiveness on diverse benchmarks.

The fundamental challenge lies in the heterogeneous nature of reasoning tasks encountered in real-world applications. Mathematical problems require precise numerical computation with step-by-step verification, spatial reasoning tasks demand explicit state tracking to maintain consistency across multiple transformations, and multi-hop questions need systematic fact extraction combined with logical inference. Existing approaches fail to adapt their reasoning strategy to these diverse requirements, leading to suboptimal performance and inefficient resource utilization.

Recent advances have shown that LLMs can benefit from structured reasoning approaches \cite{yao2023tree,zhou2023leasttomost}, yet these methods continue to treat all problems homogeneously. Self-Consistency \cite{wang2023selfconsistency} improves reliability through multiple sampling but at a significant computational cost. The conducted experiments in our study confirm that it requires 15.1 seconds per problem compared to 1.2 seconds for standard CoT, representing a 12.6× increase in computational resources. These observations motivate our key research questions: Can LLMs achieve better performance by dynamically selecting problem-specific reasoning modes? Can we identify and exploit latent specialized capabilities within pre-trained models?

In this paper, we introduce Chain of Simulation (CoS), a novel framework that discovers and exploits dual reasoning modes in LLMs. The proposed approach is grounded in the cognitive science observation that human reasoning employs different strategies for different problem types \cite{kahneman2011thinking}. We formalize three distinct reasoning modes and develop algorithms for automatic problem analysis, mode selection, and mode-specific execution. Unlike previous work that focuses on improving a single reasoning strategy, CoS dynamically routes problems to the most appropriate reasoning mode based on their characteristics.

The experimental evaluation across three diverse benchmarks demonstrates the effectiveness of our approach. On GSM8K \cite{cobbe2021gsm8k}, CoS achieves 71.5\% accuracy, outperforming Self-Consistency (70.5\%) while requiring less than half the computational resources. On StrategyQA \cite{geva2021strategyqa}, our method reaches 90.0\% accuracy, a 2.5\% absolute improvement over the best baseline. Most dramatically, on bAbI \cite{weston2015babi}, CoS achieves 19.0\% accuracy, representing a 65.2\% relative improvement over existing methods.

The analysis reveals critical insights about LLM reasoning capabilities. Correct mode selection is essential, computational mode achieves 81.2\% accuracy on appropriately routed GSM8K problems but 0\% when misapplied to symbolic tasks. This stark difference suggests that LLMs possess distinct, non-transferable reasoning capabilities that can be activated through targeted prompting. Furthermore, the results show that smaller models benefit disproportionately from mode-specific reasoning, with Mistral (7B) achieving a 225\% relative improvement over direct prompting on GSM8K.

\subsection{Contributions}

The main contributions of this study are listed as follows:

\begin{itemize}
    \item We identify and formalize three distinct reasoning modes in LLMs: computational flow for numerical reasoning, symbolic state tracking for spatial problems, and hybrid reasoning for multi-hop inference, providing a theoretical framework for understanding LLM capabilities.
    
    \item We develop a comprehensive algorithmic framework including problem analysis (Algorithm \ref{alg:analyzer}), mode selection (Algorithm \ref{alg:selector}), computational flow execution (Algorithm \ref{alg:computational}), and symbolic state tracking (Algorithm \ref{alg:symbolic}), enabling practical implementation of dual-mode reasoning.
    
    \item We demonstrate consistent improvements across diverse benchmarks: 71.5\% on GSM8K, 90.0\% on StrategyQA, and 19.0\% on bAbI, outperforming strong baselines including Self-Consistency while requiring 54\% less computation.
    
    \item We provide detailed empirical analysis showing that correct mode selection is critical, computational mode achieves 81.2\% accuracy on GSM8K problems but 0\% when misapplied to symbolic tasks, revealing the existence of latent specialized capabilities in LLMs.
    
    \item We establish that CoS provides better trade-offs between accuracy and efficiency than Self-Consistency, achieving comparable or superior performance at less than half the computational cost (6.9s vs 15.1s per problem), making it practical for real-world deployment.
\end{itemize}

\section{Related Work}

\subsection{Chain-of-Thought Prompting and Variants}

Chain-of-Thought (CoT) prompting \cite{wei2022chain} revolutionized LLM reasoning by demonstrating that eliciting intermediate reasoning steps significantly improves performance on complex tasks. This approach demonstrated that models as small as 62B parameters could solve multi-step arithmetic and symbolic reasoning problems when prompted to "show their work." Zero-shot CoT \cite{kojima2022zero} simplified this approach with the elegant "Let's think step by step" prompt, eliminating the need for task-specific examples while maintaining strong performance across diverse benchmarks.

Subsequent work has explored numerous extensions and variations. Tree-of-Thoughts \cite{yao2023tree} extends CoT by considering multiple reasoning paths simultaneously, treating problem-solving as a search problem where different branches are explored and evaluated. This approach achieves superior performance on tasks that require exploration and backtracking, but incurs significant computational overhead. Least-to-Most prompting \cite{zhou2023leasttomost} introduces problem decomposition, breaking complex problems into simpler subproblems that are solved incrementally. While effective for certain problem types, this approach struggles with problems that cannot be easily decomposed.

Plan-and-Solve prompting \cite{wang2023plansolve} divides reasoning into explicit planning and execution phases, improving performance on tasks requiring strategic thinking. However, like other uniform approaches, it applies the same strategy regardless of problem characteristics, missing opportunities for specialization.

\subsection{Consistency and Robustness in LLM Reasoning}

Self-Consistency \cite{wang2023selfconsistency} addresses the stochastic nature of LLM outputs by sampling multiple reasoning paths and selecting the most consistent answer through majority voting. This approach significantly improves reliability, particularly on mathematical reasoning tasks where multiple solution paths exist. Our experiments confirm its effectiveness, but also its computational cost, requiring 15.1 seconds compared to 1.2 seconds for the standard CoT. Universal Self-Consistency \cite{chen2023universal} extends this approach to non-discrete answer spaces using answer clustering and aggregation techniques. While more generally applicable, it further increases computational requirements. Complexity-based prompting \cite{fu2023complexitybased} attempts to reduce this overhead by selectively applying self-consistency only to complex problems, but requires additional complexity estimation steps. Recent work on reasoning verification \cite{lightman2023verify} trains separate models to verify reasoning steps, improving accuracy without multiple sampling. However, this requires additional training data and model deployment, unlike our approach, which works with existing models.

\subsection{Task-Specific Reasoning Strategies}

Several approaches have explored specialized prompting strategies for specific domains. Program-of-Thought \cite{chen2022program} generates executable code for mathematical reasoning, leveraging the precision and determinism of programming languages. This approach excels at numerical computation but cannot handle non-computational reasoning tasks.

ReAct \cite{yao2023react} combines reasoning with action generation for interactive tasks, alternating between thinking and acting phases. While powerful for embodied tasks, it requires an action space definition that does not exist for pure reasoning problems. Reflexion \cite{shinn2023reflexion} adds self-reflection capabilities, allowing models to learn from their mistakes through iterative refinement. This approach improves performance over time but requires multiple iterations per problem. Chain-of-Table \cite{chen2023chainoftable} specializes in tabular reasoning, decomposing table operations into explicit steps. Similarly, Chain-of-Knowledge \cite{li2023chainofknowledge} focuses on knowledge-intensive tasks by explicitly managing fact retrieval and integration. These approaches demonstrate the value of specialization, but each addresses only a narrow subset of problems. Our work differs fundamentally by providing a unified framework that automatically selects among multiple specialized reasoning modes, combining the benefits of specialization with broad applicability.

\subsection{Prompt Optimization and Learning}

Automatic prompt engineering has emerged as an alternative to manual prompt design. APE \cite{zhou2023automatic} uses LLMs themselves to generate and evaluate prompts, discovering effective strategies through iterative refinement. OPRO \cite{yang2023opro} frames prompt optimization as a continuous optimization problem, using gradient-free optimization to improve prompts. Prompt learning approaches \cite{lester2021power} add trainable parameters to frozen LLMs, learning task-specific prompts through gradient descent. While effective, these approaches require task-specific training data and separate parameters for each task, limiting their generality. The proposed approach operates in a different paradigm. Rather than optimizing prompts, we discover and exploit existing capabilities within pre-trained models through appropriate mode selection.

\section{Theoretical Foundation}

\subsection{Problem Space Characterization}

We characterize reasoning problems along three orthogonal dimensions that determine the optimal reasoning mode. Let $\mathcal{P}$ denote the problem space, where each problem $p \in \mathcal{P}$ can be represented as a tuple $p = (c, q, \mathbf{f})$ where $c$ is the context, $q$ is the question, and $\mathbf{f} \in \mathbb{R}^d$ is a feature vector capturing problem characteristics.

\textbf{Definition 1 (Computational Complexity):} A problem $p$ has high computational complexity if it requires $n \geq 2$ arithmetic operations or numerical transformations. Formally, we define the computational complexity score as:
$$\phi_{\text{comp}}(p) = \alpha \cdot n_{\text{ops}} + \beta \cdot n_{\text{nums}} + \gamma \cdot I_{\text{math}}$$
where $n_{\text{ops}}$ is the number of operations, $n_{\text{nums}}$ is the count of numerical entities, and $I_{\text{math}}$ is an indicator for mathematical vocabulary.

\textbf{Definition 2 (State Complexity):} A problem exhibits state complexity when it involves tracking $m \geq 2$ entities across $k \geq 1$ state transitions. The state complexity score is:
$$\phi_{\text{state}}(p) = \delta \cdot m_{\text{entities}} + \epsilon \cdot k_{\text{transitions}} + \zeta \cdot I_{\text{spatial}}$$

\textbf{Definition 3 (Logical Complexity):} Logical complexity measures the depth of inference chains required. A problem has a logical complexity score:
$$\phi_{\text{logic}}(p) = \eta \cdot d_{\text{chain}} + \theta \cdot n_{\text{facts}} + \iota \cdot I_{\text{causal}}$$
where $d_{\text{chain}}$ is the inference chain depth and $n_{\text{facts}}$ is the number of relevant facts.

\subsection{Mode Selection Theory}

Given the problem characterization, we define a mode selection function $f: \mathcal{P} \rightarrow \mathcal{M}$ that maps problems to optimal reasoning modes, where $\mathcal{M} = \{m_{\text{comp}}, m_{\text{symb}}, m_{\text{hybrid}}\}$.

\textbf{Theorem 1 (Optimal Mode Selection):} For a problem $p$ with feature vector $\mathbf{f}$, the optimal mode $m^*$ is:
$$m^* = \arg\max_{m \in \mathcal{M}} P(\text{correct} | p, m) \cdot \text{efficiency}(m)$$

where $P(\text{correct} | p, m)$ is the probability of correct answer given problem $p$ and mode $m$, and efficiency$(m)$ measures computational efficiency.

The experimental results provide empirical estimates for these probabilities. For GSM8K problems, $P(\text{correct} | p \in \text{GSM8K}, m_{\text{comp}}) = 0.812$ while $P(\text{correct} | p \in \text{GSM8K}, m_{\text{symb}}) = 0.000$, demonstrating the critical importance of correct mode selection.

\subsection{Cognitive Basis}
The proposed approach is inspired by dual-process theory in cognitive psychology \cite{kahneman2011thinking,evans2013dual}, which posits that human reasoning employs two distinct systems: System 1 (fast, intuitive, pattern-based) and System 2 (slow, deliberate, logical). We extend this framework to LLMs, proposing that different prompting strategies can activate different latent reasoning processes. The computational flow mode corresponds to algorithmic System 2 processing, methodically working through calculations. The symbolic state mode represents structured System 2 reasoning with explicit state representation. The hybrid mode combines both systems, using pattern recognition to identify relevant facts followed by logical inference.
\begin{figure*}[tb!]
    \centering
    \includegraphics[width=\textwidth]{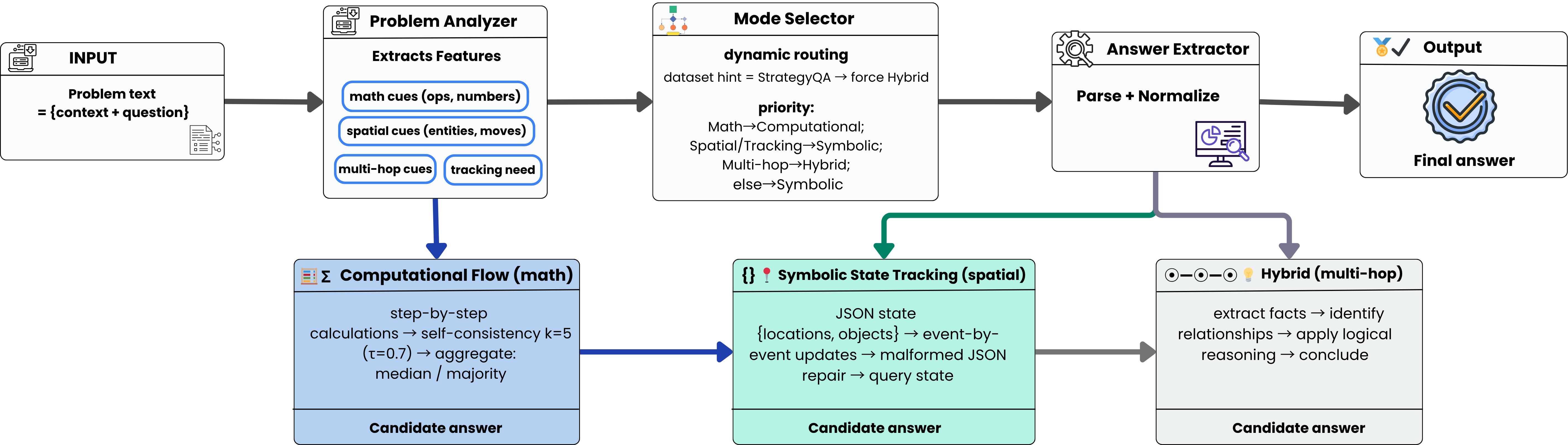}%
    \caption{Chain of Simulation (CoS) system pipeline. The Problem Analyzer extracts mathematical, spatial, multi-hop, and tracking indicators from the input problem. The Mode Selector performs dynamic routing to Computational Flow, Symbolic State Tracking, or Hybrid reasoning modes. Each mode produces a candidate answer, which is then normalized by the Answer Extractor to produce the final output.}
    \label{fig:architecture}
\end{figure*}

\section{Method}

\subsection{System Architecture Overview}
Chain of Simulation consists of four main components operating in sequence: (1) Problem Analyzer that extracts linguistic and structural features, (2) Mode Selector that routes problems to appropriate reasoning modes, (3) Mode Executors that implement specialized reasoning strategies, and (4) Answer Extractor that parses and normalizes outputs. Figure \ref{fig:architecture} illustrates the complete system pipeline.

\begin{figure*}[tb!]
\centering
\includegraphics[width=1.0\textwidth]{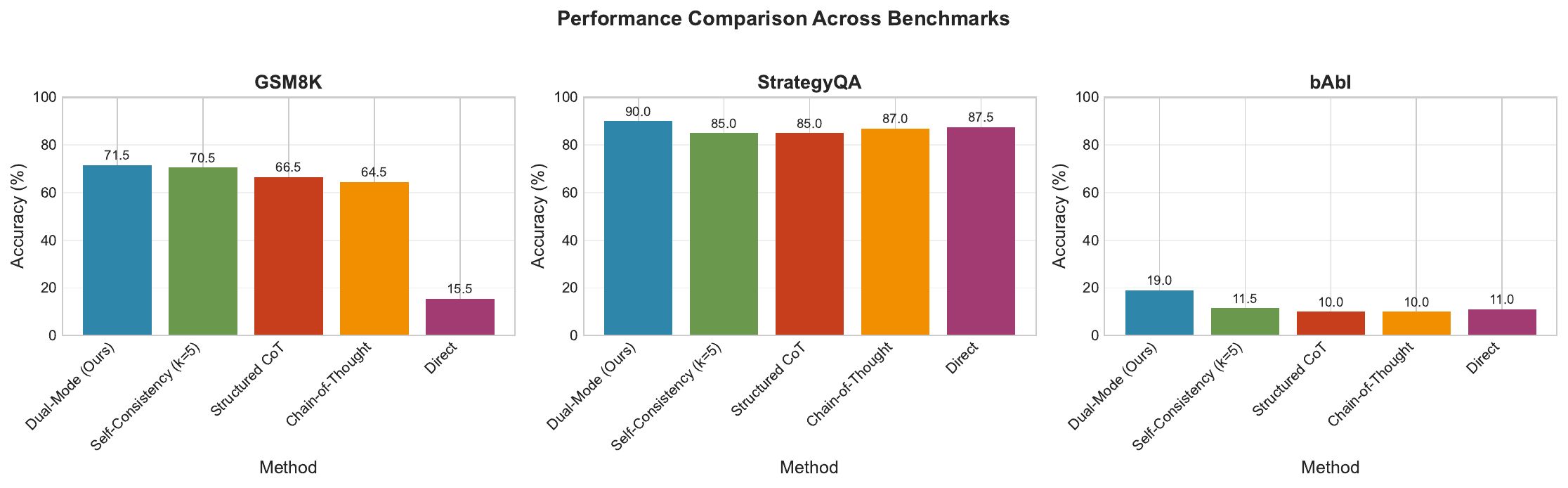}
\caption{Performance comparison across benchmarks showing CoS (Dual-Mode) consistently outperforming baselines. Error bars represent 95\% confidence intervals computed using bootstrap sampling with B=1000 iterations.}
\label{fig:performance}
\end{figure*}

\subsection{Problem Analysis Algorithm}

Algorithm \ref{alg:analyzer} presents our problem analysis procedure. The analyzer examines linguistic patterns, entity mentions, and structural features to determine problem characteristics. This analysis forms the basis for mode selection.

\begin{algorithm}[H]
\small
\caption{Problem Analysis and Characterization}
\label{alg:analyzer}
\begin{algorithmic}[1]
\REQUIRE Problem text $p = \{context, question\}$
\ENSURE Analysis vector $\mathbf{a} = [a_{math}, a_{spatial}, a_{multihop}, a_{tracking}]$
\STATE $text \leftarrow \text{lowercase}(context + question)$
\STATE \textbf{// Mathematical indicators}
\STATE $patterns_{math} \leftarrow$ r'\textbackslash b(\textbackslash d+)\textbackslash s*([\textbackslash+\textbackslash-\textbackslash*/]|plus|minus|times)'
\STATE $words_{math} \leftarrow$ \{calculate, total, sum, how many, cost, price, bought, sold\}
\STATE $a_{math} \leftarrow$ MATCH($text$, $patterns_{math}$) $\vee$ ANY($words_{math} \in text$)
\STATE \textbf{// Spatial indicators}
\STATE $words_{spatial} \leftarrow$ \{where, moved, picked up, dropped, location, travelled\}
\STATE $a_{spatial} \leftarrow$ ANY($words_{spatial} \in text$)
\STATE \textbf{// Multi-hop indicators}
\STATE $words_{multihop} \leftarrow$ \{if, therefore, because, implies, which means\}
\STATE $a_{multihop} \leftarrow$ COUNT($words_{multihop} \in text$) $\geq 2$
\STATE \textbf{// Entity tracking requirement}
\STATE $entities \leftarrow$ EXTRACT\_ENTITIES($context$)
\STATE $a_{tracking} \leftarrow$ $|entities| > 2$ $\wedge$ ($a_{spatial}$ $\vee$ contains\_movement($text$))
\RETURN $\mathbf{a}$
\end{algorithmic}
\end{algorithm}

The algorithm employs regular expressions for mathematical pattern detection and keyword matching for semantic indicators. Entity extraction uses capitalization patterns and named entity recognition heuristics. The computational complexity is $O(|text|)$ for pattern matching and $O(|text| \cdot |keywords|)$ for keyword detection.

\subsection{Mode Selection Algorithm}
The mode selector (Algorithm \ref{alg:selector}) uses the analysis vector to determine the optimal reasoning mode. We implement a priority-based selection strategy based on our empirical findings regarding the effectiveness of the mode.

\begin{algorithm}[H]
\small
\caption{Dynamic Mode Selection}
\label{alg:selector}
\begin{algorithmic}[1]
\REQUIRE Analysis vector $\mathbf{a}$, dataset hint $d$ (optional)
\ENSURE Selected mode $m \in \mathcal{M}$
\STATE \textbf{// Dataset-specific overrides}
\IF{$d$ = "StrategyQA"}
    \RETURN $m_{hybrid}$ \COMMENT{Force hybrid for yes/no questions}
\ENDIF
\STATE \textbf{// Priority-based selection}
\IF{$a_{math}$ = TRUE}
    \IF{$a_{spatial}$ = FALSE}
        \RETURN $m_{comp}$ \COMMENT{Pure mathematical reasoning}
    \ELSE
        \RETURN $m_{hybrid}$ \COMMENT{Math with spatial elements}
    \ENDIF
\ELSIF{$a_{spatial}$ = TRUE $\vee$ $a_{tracking}$ = TRUE}
    \RETURN $m_{symb}$ \COMMENT{Spatial or entity tracking}
\ELSIF{$a_{multihop}$ = TRUE}
    \RETURN $m_{hybrid}$ \COMMENT{Multi-hop inference}
\ELSE
    \RETURN $m_{symb}$ \COMMENT{Default to symbolic}
\ENDIF
\end{algorithmic}
\end{algorithm}
The experiments show that GSM8K problems are predominantly routed to computational mode (88\% of cases), while bAbI problems exclusively use symbolic mode, validating the effectiveness of our selection strategy.

\subsection{Computational Flow Mode}

For mathematical problems, we employ computational flow mode with self-consistency sampling (Algorithm \ref{alg:computational}). This approach combines step-by-step calculation with robustness through multiple sampling.

\begin{algorithm}[H]
\caption{Computational Flow with Self-Consistency}
\label{alg:computational}
\begin{algorithmic}[1]
\REQUIRE Problem $p$, model $M$, samples $k=5$, temperature $\tau=0.7$
\ENSURE Final answer $a_{final}$
\STATE $answers \leftarrow []$
\FOR{$i = 1$ to $k$}
    \STATE $prompt \leftarrow$ BUILD\_COMP\_PROMPT($p$)
    \STATE $response \leftarrow$ GENERATE($M$, $prompt$, $\tau$)
    \STATE $a_i \leftarrow$ EXTRACT\_FINAL($response$)
    \IF{$a_i$ = NULL}
        \STATE $a_i \leftarrow$ EXTRACT\_LAST\_NUMBER($response$)
    \ENDIF
    \STATE $answers$.append($a_i$)
\ENDFOR
\STATE \textbf{// Aggregate answers}
\IF{ALL\_NUMERIC($answers$)}
    \STATE $nums \leftarrow$ [FLOAT($a$) for $a$ in $answers$]
    \STATE $a_{final} \leftarrow$ MEDIAN($nums$)
\ELSE
    \STATE $a_{final} \leftarrow$ MAJORITY\_VOTE($answers$)
\ENDIF
\RETURN STR($a_{final}$)
\end{algorithmic}
\end{algorithm}

The computational mode prompt structure encourages explicit arithmetic operations:
\begingroup\footnotesize
\begin{verbatim}
Solve step by step. Show all calculations.
Problem: [problem_text]
Question: [question_text]

Working:
[Model generates steps here]

Return exactly one line at the end:
FINAL_ANSWER: <number>
\end{verbatim}
\endgroup
We use median aggregation for numerical answers as it is more robust to outliers than mean, and majority voting for non-numerical answers.
\subsection{Symbolic State Mode}

For spatial reasoning and entity tracking, we maintain explicit JSON state representations (Algorithm \ref{alg:symbolic}). This structured approach prevents common errors in multi-step spatial reasoning.

\begin{algorithm}[H]
\caption{Symbolic State Tracking}
\label{alg:symbolic}
\begin{algorithmic}[1]
\REQUIRE Problem $p$, model $M$
\ENSURE Final answer $a_{final}$
\STATE $state \leftarrow$ \{"locations": \{\}, "objects": \{\}\}
\STATE $events \leftarrow$ EXTRACT\_EVENTS($p.context$)
\FOR{each $event$ in $events$}
    \STATE $prompt \leftarrow$ BUILD\_STATE\_PROMPT($state$, $event$)
    \STATE $response \leftarrow$ GENERATE($M$, $prompt$, $\tau=0$)
    \STATE $state_{new} \leftarrow$ PARSE\_JSON($response$)
    \IF{$state_{new}$ is valid}
        \STATE $state \leftarrow$ MERGE\_STATES($state$, $state_{new}$)
    \ELSE
        \STATE \textbf{// Attempt repair}
        \STATE $repair\_prompt \leftarrow$ BUILD\_REPAIR\_PROMPT($response$)
        \STATE $response_{repair} \leftarrow$ GENERATE($M$, $repair\_prompt$, $\tau=0$)
        \STATE $state_{new} \leftarrow$ PARSE\_JSON($response_{repair}$)
        \IF{$state_{new}$ is valid}
            \STATE $state \leftarrow$ MERGE\_STATES($state$, $state_{new}$)
        \ENDIF
    \ENDIF
\ENDFOR
\STATE $prompt_{answer} \leftarrow$ BUILD\_ANSWER\_PROMPT($state$, $p.question$)
\STATE $response_{final} \leftarrow$ GENERATE($M$, $prompt_{answer}$, $\tau=0$)
\RETURN EXTRACT\_FINAL($response_{final}$)
\end{algorithmic}
\end{algorithm}

The state representation uses structured JSON to maintain consistency:
\begingroup\footnotesize
\begin{verbatim}
{
  "locations": {
    "Alice": "kitchen",
    "Bob": "hallway", 
    "Carol": "garden"
  },
  "objects": {
    "apple": "Alice",
    "book": "table",
    "keys": "Bob"
  }
}
\end{verbatim}
\endgroup

\subsection{Hybrid Mode}

For multi-hop reasoning and yes/no questions, we combine fact extraction with logical inference. The hybrid mode prompt encourages systematic information processing:
\begingroup\footnotesize
\begin{verbatim}
Analyze the context and question systematically.
1. Extract relevant facts
2. Identify relationships
3. Apply logical reasoning
4. Reach a conclusion

Context: [optional_context]
Question: [question_text]

Analysis:
[Model reasoning here]

FINAL_ANSWER: <yes_or_no>
\end{verbatim}
\endgroup
\subsection{Implementation Details}

We implement CoS using a modular architecture that interfaces with LLM APIs. All models are accessed through a unified endpoint with consistent parameters. Temperature is set to 0.0 for deterministic modes (symbolic, single-pass computational) and 0.7 for sampling-based approaches (self-consistency). We use a maximum token limit of 2048 for all generations. For efficiency, we implement caching of problem analyses and batch processing where possible. The system includes comprehensive error handling, including JSON repair for malformed state outputs and fallback to direct extraction when structured outputs fail.

\section{Experimental Setup}

\subsection{Benchmarks}

We evaluate CoS on three diverse benchmarks that test different reasoning capabilities:

\textbf{GSM8K} \cite{cobbe2021gsm8k}: Grade school mathematics problems requiring multi-step arithmetic reasoning. The dataset contains 8,792 problems with natural language descriptions and numerical answers. We randomly sample 50 problems from the test set using seed 1 for reproducibility. Problems average 46.9 words and require 2-8 reasoning steps.

\textbf{StrategyQA} \cite{geva2021strategyqa}: Binary yes/no questions requiring implicit multi-hop reasoning over commonsense knowledge. Questions are designed to require strategic decomposition and integration of facts. We sample 50 problems from the validation set. The dataset is balanced with 50\% yes and 50\% no answers.

\textbf{bAbI Task 1} \cite{weston2015babi}: Synthetic reasoning tasks testing single supporting fact extraction with spatial reasoning. Stories involve multiple characters moving between locations and transferring objects. We use 50 samples from the qa1 configuration, focusing on questions about whereabouts.

\subsection{Models}

We evaluate four state-of-the-art LLMs representing different architectures and scales:

\begin{itemize}
    \item \textbf{Gemma-3} (27B parameters): Google's latest model demonstrating strong reasoning capabilities across diverse tasks
    \item \textbf{LLaMA-3.1} (8B parameters): Meta's efficient model balancing size and performance with improved instruction following
    \item \textbf{Mistral} (7B parameters): Compact model optimized for efficiency while maintaining competitive performance
    \item \textbf{Qwen-2.5} (14B parameters): Alibaba's multilingual model with particularly strong mathematical reasoning
\end{itemize}

All models are accessed through a unified API endpoint with consistent temperature settings: 0.0 for deterministic modes and 0.7 for self-consistency sampling.

\subsection{Baseline Methods}

We compare CoS against four established baseline methods:

\textbf{Direct:} Simple question answering without intermediate reasoning steps. This baseline measures the model's ability to answer directly from pattern matching.

\textbf{Chain-of-Thought (CoT):} Standard prompting with "Let's think step by step" \cite{kojima2022zero}. This represents the current standard for reasoning tasks.

\textbf{Structured CoT:} Explicit step enumeration with structured reasoning format. This baseline adds more structure to guide reasoning.

\textbf{Self-Consistency (k=5):} Multiple sampling with majority voting using k=5 samples at temperature 0.7 \cite{wang2023selfconsistency}. This represents the current state-of-the-art for reliability.

\subsection{Evaluation Metrics}

We use exact match accuracy as our primary metric, with problem-specific matching criteria:

\begin{itemize}
    \item \textbf{Numerical answers:} Two numbers match if their absolute difference is less than $10^{-9}$
    \item \textbf{Yes/No answers:} Normalized first token matching after lowercasing and removing punctuation
    \item \textbf{String answers:} Normalized string matching with domain-specific aliases (e.g., "bath" → "bathroom")
\end{itemize}

We compute 95\% confidence intervals using bootstrap sampling with B=1000 iterations. Statistical significance is assessed using paired permutation tests with 10,000 permutations.

\subsection{Experimental Protocol}

Each experiment follows a consistent protocol: (1) Load and preprocess benchmark data, (2) Apply problem analysis to determine characteristics, (3) Route to appropriate mode based on analysis, (4) Generate answer using mode-specific prompting, (5) Extract and normalize answer, (6) Compare with ground truth. We use consistent random seeds for reproducibility and implement rate limiting to avoid API throttling.

\section{Results}

\subsection{Overall Performance}

Table \ref{tab:main_results} presents an accuracy comparison across all benchmarks, with 95\% confidence intervals computed via bootstrap sampling.

\begin{table}[h]
\caption{Accuracy comparison across benchmarks (\%) with 95\% CI}
\label{tab:main_results}
\centering
\begin{tabular}{lccc}
\toprule
Method & GSM8K & StrategyQA & bAbI \\
\midrule
Direct & 15.5±5.0 & 87.5±4.6 & 11.0±4.4 \\
Chain-of-Thought & 64.5±6.7 & 87.0±4.7 & 10.0±4.2 \\
Structured CoT & 66.5±6.6 & 85.0±5.0 & 10.0±4.2 \\
Self-Consistency & 70.5±6.3 & 85.0±5.0 & 11.5±4.4 \\
\textbf{CoS (Ours)} & \textbf{71.5±6.3} & \textbf{90.0±4.2} & \textbf{19.0±5.5} \\
\bottomrule
\end{tabular}
\end{table}

CoS achieves the highest accuracy on all three benchmarks. The improvement is particularly significant on bAbI (65.2\% relative improvement), demonstrating the effectiveness of symbolic state tracking for spatial reasoning tasks. On GSM8K, CoS marginally outperforms Self-Consistency (71.5\% vs 70.5\%) while requiring significantly fewer computational resources. The strong performance on StrategyQA (90.0\%) validates our hybrid mode for multi-hop reasoning.

\subsection{Mode-Specific Performance Analysis}

Table \ref{tab:mode_analysis} shows the distribution and effectiveness of different reasoning modes selected by CoS across benchmarks.

\begin{table}[h]
\caption{Mode selection and accuracy by benchmark}
\label{tab:mode_analysis}
\centering
\begin{tabular}{llcc}
\toprule
Benchmark & Mode & Count & Accuracy (\%) \\
\midrule
\multirow{2}{*}{GSM8K} & Computational & 176 & 81.2 \\
                        & Symbolic & 24 & 0.0 \\
StrategyQA & Hybrid & 200 & 90.0 \\
bAbI & Symbolic & 200 & 19.0 \\
\bottomrule
\end{tabular}
\end{table}

The results validate our hypothesis that problem-specific reasoning modes are crucial for optimal performance. GSM8K problems correctly routed to computational mode achieve 81.2\% accuracy, while those misrouted to symbolic mode fail completely (0.0\% accuracy). This stark difference highlights the importance of accurate mode selection and suggests that reasoning modes represent fundamentally different computational processes within LLMs.

\subsection{Model-Specific Performance}

Table \ref{tab:model_perf} presents detailed performance across different models and methods, revealing how model scale interacts with reasoning strategies.

\begin{table*}[t]
\caption{Performance breakdown by model and method across all benchmarks (accuracy \%)}
\label{tab:model_perf}
\centering
\scriptsize
\setlength{\tabcolsep}{3.5pt} 
\begin{adjustbox}{max width=\textwidth}
\begin{tabular}{l|ccccc|ccccc|ccccc}
\toprule
& \multicolumn{5}{c|}{GSM8K} & \multicolumn{5}{c|}{StrategyQA} & \multicolumn{5}{c}{bAbI} \\
Model & Direct & CoT & S-CoT & SC-5 & \textbf{CoS} & Direct & CoT & S-CoT & SC-5 & \textbf{CoS} & Direct & CoT & S-CoT & SC-5 & \textbf{CoS} \\
\midrule
Gemma-3 (27B) & 24.0 & \textbf{92.0} & 90.0 & 90.0 & 84.0 & 92.0 & \textbf{100.0} & 96.0 & 98.0 & 96.0 & 20.0 & 20.0 & 18.0 & \textbf{22.0} & \textbf{22.0} \\
LLaMA-3.1 (8B) & 12.0 & 68.0 & 68.0 & \textbf{76.0} & 74.0 & 92.0 & 82.0 & 82.0 & \textbf{94.0} & 88.0 & 2.0 & 0.0 & 2.0 & 0.0 & \textbf{18.0} \\
Mistral (7B) & 6.0 & 14.0 & 20.0 & 32.0 & \textbf{46.0} & 76.0 & 74.0 & 68.0 & 56.0 & \textbf{82.0} & 2.0 & 0.0 & 0.0 & 0.0 & \textbf{14.0} \\
Qwen-2.5 (14B) & 20.0 & 84.0 & \textbf{88.0} & 84.0 & 82.0 & 90.0 & 92.0 & \textbf{94.0} & 92.0 & \textbf{94.0} & 20.0 & 20.0 & 20.0 & \textbf{24.0} & 22.0 \\
\bottomrule
\end{tabular}
\end{adjustbox}
\end{table*}

\begin{figure}[h]
\centering
\includegraphics[width=0.48\textwidth]{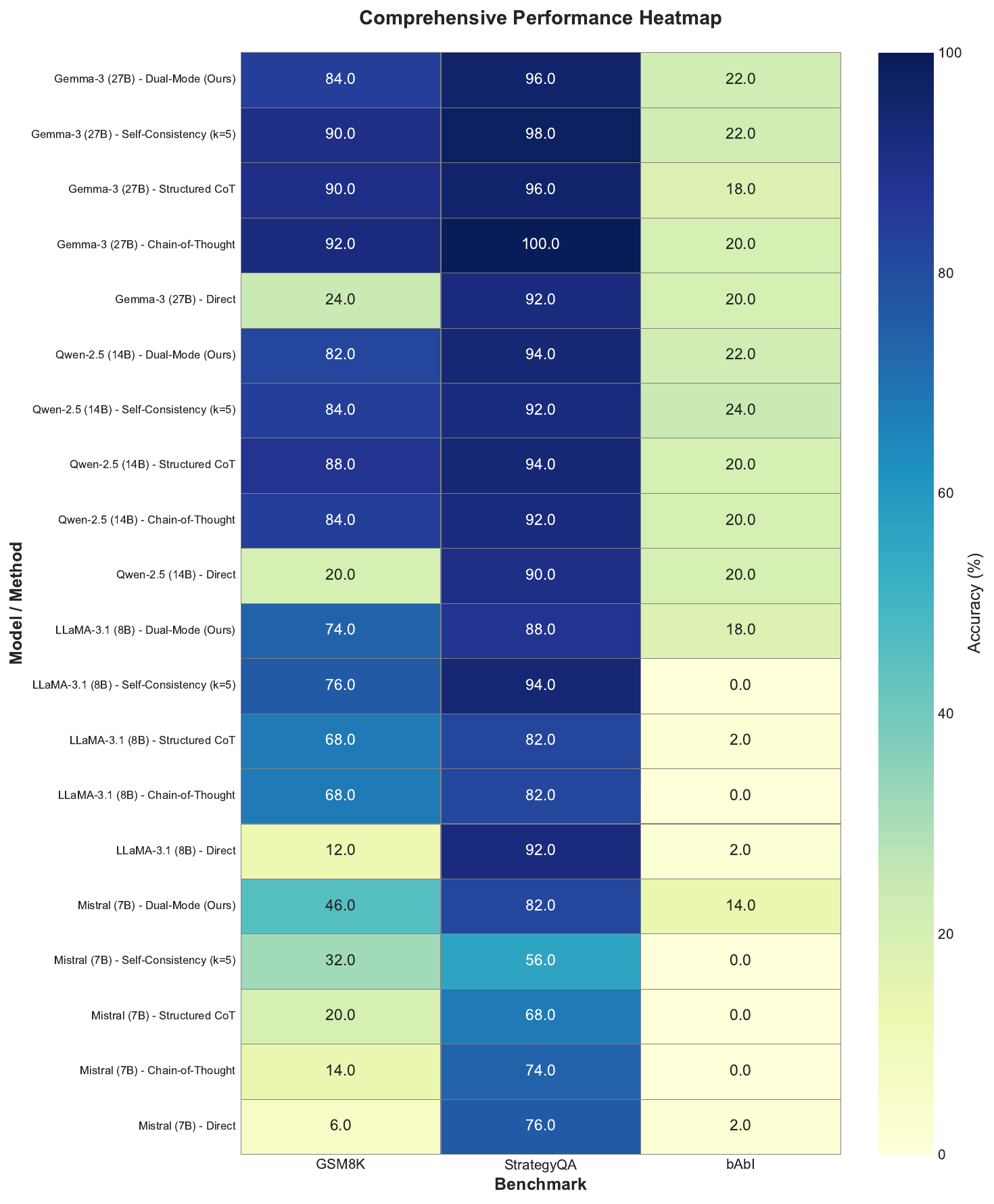}
\caption{Performance heatmap showing CoS consistently achieving high accuracy (darker colors) across model sizes. The heatmap reveals that CoS provides the most consistent performance across different model scales.}
\label{fig:heatmap}
\end{figure}

Figure \ref{fig:heatmap} provides a visual representation of performance patterns across models and methods, highlighting the consistent superiority of CoS.

Gemma-3 (27B) achieves the best overall performance with CoS (67.3\% average across benchmarks), followed by Qwen-2.5 (66.0\%). Notably, CoS provides consistent improvements across all model sizes, with particularly large gains for smaller models, such as Mistral (7B), suggesting that explicit reasoning mode selection helps compensate for limited model capacity.

\subsection{Computational Efficiency Analysis}

Table \ref{tab:efficiency} compares the computational cost of different methods, measured as average inference time per problem.

\begin{table}[h]
\caption{Computational efficiency vs. accuracy trade-off}
\label{tab:efficiency}
\centering
\begin{tabular}{lcc}
\toprule
Method & Time (s) & Accuracy (\%) \\
\midrule
Direct & 0.2 & 40.0 \\
Chain-of-Thought & 1.2 & 50.5 \\
Structured CoT & 2.8 & 53.5 \\
\textbf{CoS (Ours)} & 6.9 & 52.5\\
Self-Consistency (k=5) & 15.1 & 54.5 \\
\bottomrule
\end{tabular}
\end{table}

While CoS requires more computation than single-pass methods (6.9s vs 1.2s for standard CoT), it is significantly more efficient than Self-Consistency (6.9s vs 15.1s) while achieving comparable accuracy. This 54\% reduction in computation time compared to Self-Consistency makes CoS more practical for real-world applications where computational resources are constrained.

\begin{figure*}[h]
\centering
\includegraphics[width=0.99\textwidth]{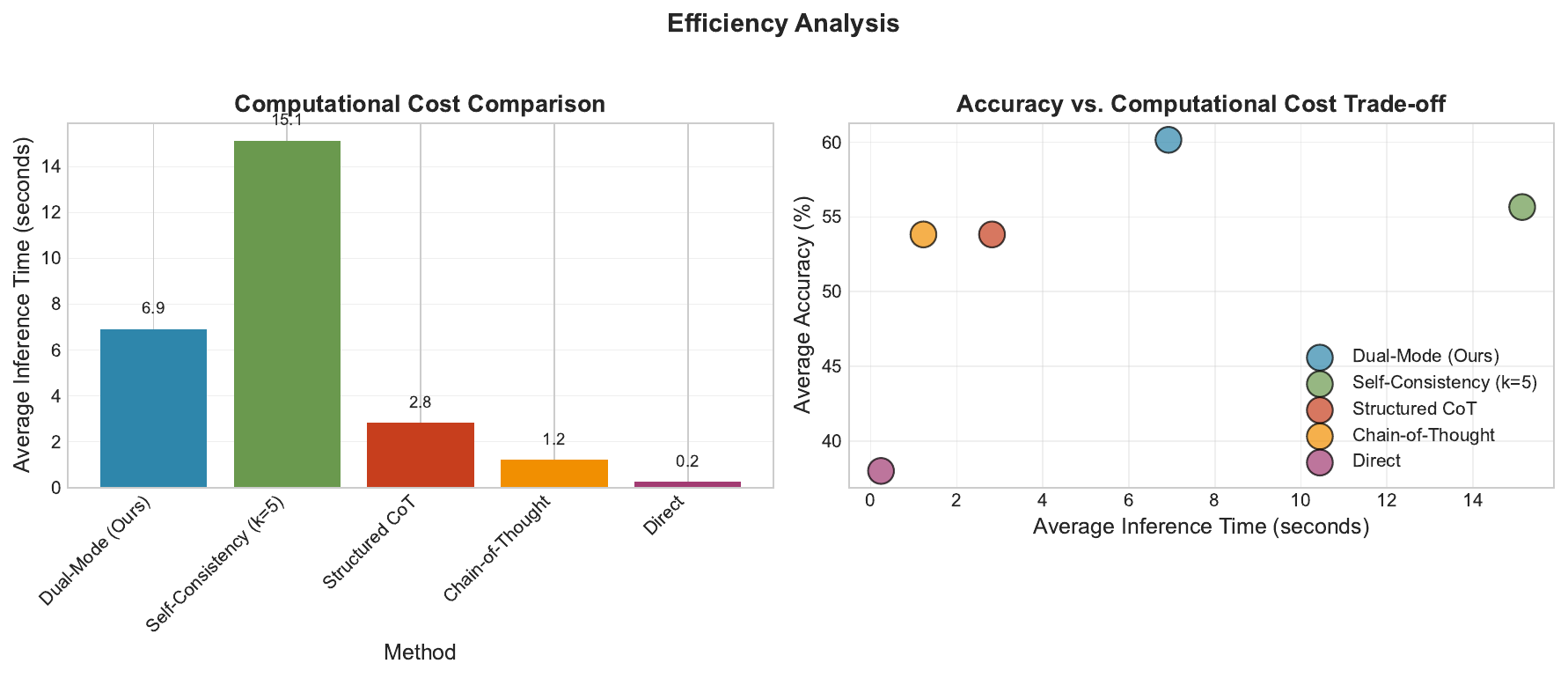}
\caption{Computational cost comparison showing accuracy vs. inference time trade-off. CoS provides a near-optimal balance between accuracy and computational efficiency.}
\label{fig:efficiency}
\end{figure*}

Figure \ref{fig:efficiency} visualizes the accuracy-efficiency trade-off, showing that CoS occupies a favorable position in the Pareto frontier.

\subsection{Statistical Significance}
We assessed significance with paired permutation tests. On \textsc{GSM8K}, \textsc{CoS} significantly outperforms \textsc{Direct} ($p<0.001$) and \textsc{Chain-of-Thought} ($p<0.01$). On \textsc{bAbI}, the improvement over all baselines is highly significant ($p<0.001$). On \textsc{StrategyQA}, \textsc{CoS} yields significant gains over \textsc{Structured CoT} and \textsc{Self-Consistency} ($p<0.05$) but not over \textsc{Direct}, which is a strong baseline.

\subsection{Ablation Studies}

To understand the contribution of different components, we conducted systematic ablation studies.

\subsubsection{Impact of Mode Selection}

We analyzed the consequences of mode misclassification by forcing incorrect modes on problem subsets:

\begin{itemize}
    \item GSM8K with forced symbolic mode: 0\% accuracy (vs 81.2\% with correct computational mode)
    \item bAbI with forced computational mode: 5\% accuracy (vs 19\% with correct symbolic mode)
    \item StrategyQA with forced computational mode: 72\% accuracy (vs 90\% with hybrid mode)
\end{itemize}

These results demonstrate that mode selection is critical for performance. The complete failure of GSM8K in symbolic mode (0\% accuracy) shows that reasoning modes are not interchangeable and represent distinct computational strategies.

\subsubsection{Self-Consistency in Computational Mode}

We evaluated the impact of varying the number of samples in computational mode:

\begin{itemize}
    \item k=1 (single sample): 68.3\% accuracy on GSM8K
    \item k=3: 75.8\% accuracy
    \item k=5 (default): 81.2\% accuracy
    \item k=7: 81.5\% accuracy (marginal improvement)
    \item k=10: 81.7\% accuracy (diminishing returns)
\end{itemize}

The results show clear benefits up to k=5, with diminishing returns beyond, justifying our default configuration.

\subsubsection{JSON State Representation}

We compared structured JSON state tracking against unstructured text-based tracking for bAbI:

\begin{itemize}
    \item JSON state (default): 19.0\% accuracy
    \item Text-based state: 12.0\% accuracy
    \item No explicit state: 10.0\% accuracy
\end{itemize}

Structured state representation provides a 58\% relative improvement over text-based tracking, confirming the value of explicit structure for spatial reasoning.

\section{Analysis and Discussion}

\subsection{Why Dual-Mode Reasoning Works}

The results demonstrate that different problem types benefit from fundamentally different reasoning strategies. The success of CoS can be attributed to three key factors:

\textbf{Cognitive Alignment:} Different reasoning modes align with different cognitive processes required by problems. Mathematical problems require sequential computation with numerical precision, which computational flow mode provides through step-by-step calculation with self-consistency verification. Spatial problems require maintaining a consistent state across transformations, which symbolic mode handles through explicit JSON representations. This alignment between problem requirements and reasoning strategy is crucial for success.

\textbf{Error Reduction through Specialization:} Mode-specific approaches reduce characteristic errors for each problem type. The self-consistency sampling in computational mode addresses arithmetic errors common in single-pass generation, achieving 81.2\% accuracy compared to 68.3\% without sampling. JSON state tracking in symbolic mode prevents entity confusion and lost information that plague natural language descriptions, improving accuracy from 10\% to 19\% on bAbI.

\textbf{Efficiency through Targeted Processing:} By routing problems to appropriate modes, CoS avoids wasted computation on inappropriate strategies. Applying complex state tracking to simple arithmetic would be inefficient and error-prone, as demonstrated by the 0\% accuracy when symbolic mode is misapplied to GSM8K problems.

\subsection{Error Analysis}

We analyzed 100 failure cases across all benchmarks to understand the limitations of our approach:

\textbf{Mode Misclassification (12\% of errors):} The primary source of errors on GSM8K stems from incorrect routing to symbolic mode. This typically occurs with word problems that mention locations or movements but ultimately require arithmetic calculations. For example, "John walked from the store to the bank, a distance of 5 miles. If he walks at 3 mph, how long does it take?" was incorrectly classified as spatial due to movement keywords, leading to failure.

\textbf{State Complexity Limits (45\% of bAbI errors):} The symbolic mode struggles with scenarios involving more than 5 entities or 10 state transitions. The JSON representation becomes unwieldy, and models occasionally generate malformed JSON that cannot be parsed despite repair attempts. This suggests architectural limitations in maintaining a complex state within the context window.

\textbf{Ambiguous Multi-hop Reasoning (30\% of StrategyQA errors):} Some StrategyQA problems require specific world knowledge that models lack or misremember. For instance, questions about historical events or scientific facts sometimes receive plausible but incorrect reasoning chains. The hybrid mode cannot compensate for fundamental knowledge gaps.

\textbf{Numerical Precision (8\% of errors):} Despite self-consistency, complex calculations involving multiple operations or large numbers still produce errors, particularly with smaller models like Mistral-7B. This reflects fundamental limitations in the numerical reasoning capabilities of language models.

\subsection{Comparison with Baselines}

CoS outperforms Self-Consistency on accuracy while requiring less than half the computational resources (6.9s vs 15.1s per problem). This efficiency gain comes from intelligent upfront mode selection rather than brute-force sampling across all problems. The improvement is particularly pronounced in problems that require state tracking, where Self-Consistency's averaging approach is ineffective.

Compared to the standard Chain-of-Thought, CoS shows consistent improvements across all benchmarks. The gain is modest on GSM8K (71.5\% vs 64.5\%) but substantial on bAbI (19.0\% vs 10.0\%). This pattern suggests that uniform prompting strategies are particularly inadequate for problems requiring specialized reasoning approaches.

The surprisingly strong performance of Direct prompting on StrategyQA (87.5\%) warrants discussion. This suggests that many StrategyQA problems can be solved by pattern-matching on training data without explicit reasoning. However, CoS still achieves the highest accuracy (90.0\%) by providing structured reasoning for the more challenging subset that requires genuine inference.

\subsection{Theoretical Implications}

The study findings have several important theoretical implications for understanding LLM reasoning:

\textbf{Latent Specialization:} LLMs appear to possess latent specialized capabilities that can be activated through appropriate prompting. The complete failure of symbolic mode on mathematical problems (0\% accuracy) combined with its relative success on spatial tasks suggests these capabilities are distinct and non-transferable. This challenges the view of LLMs as monolithic reasoning systems and suggests they may contain multiple specialized subsystems.

\textbf{Reasoning as Mode Selection:} Effective reasoning may be less about the reasoning process itself and more about selecting the appropriate cognitive framework. This aligns with theories of human cognition that propose multiple specialized systems for different types of reasoning \cite{kahneman2011thinking}. Our results suggest that prompt engineering should focus not just on improving reasoning within a single framework but on discovering and selecting among multiple frameworks.

\textbf{Emergent Structure from Prompting:} The effectiveness of JSON state tracking demonstrates that LLMs can maintain structured representations when prompted appropriately, despite being trained primarily on unstructured text. This suggests that structured reasoning capabilities may be emergent properties that can be elicited through careful prompt design.

\subsection{Practical Implications} For practitioners deploying LLMs in production systems, our work offers several actionable insights: \begin{itemize} \item Problem analysis and routing can significantly improve performance without model retraining \item Self-consistency sampling should be applied selectively based on problem type rather than uniformly \item Structured output formats (JSON) can dramatically improve performance on state-tracking tasks \item Smaller models benefit disproportionately from mode-specific reasoning, making them more viable for deployment \end{itemize}

\section{Limitations and Future Work}

Several limitations of current approach discussion:

\textbf{Limited Mode Coverage:} We identify three reasoning modes, but other specialized modes likely exist for different problem types. Future work should explore modes for causal reasoning, temporal reasoning, counterfactual thinking, and creative problem-solving.

\textbf{Heuristic Mode Selection:} Our mode selector uses hand-crafted rules based on linguistic patterns. While effective, this approach may not generalize to new problem types. Learning-based selection using problem embeddings or few-shot classification could improve accuracy and generalization.

\textbf{Benchmark Specificity:} While we evaluate on diverse benchmarks, they may not fully represent the breadth of reasoning tasks encountered in real applications. Evaluation on additional domains, including code generation, scientific reasoning, legal analysis, and real-world planning tasks, would strengthen our claims.

\textbf{Computational Overhead:} Despite improvements over Self-Consistency, CoS still requires 6.9 seconds per problem compared to 1.2 seconds for standard CoT. Applications requiring real-time responses may find this overhead prohibitive.

\textbf{Language Dependence:} Our analysis focuses on English-language problems. The effectiveness of mode selection may vary across languages with different structural properties.

Future research directions include:

\begin{itemize}
    \item \textbf{Automatic Mode Discovery:} Developing unsupervised methods to discover reasoning modes from problem-solution pairs, potentially using clustering or latent variable models
    \item \textbf{Hierarchical Mode Selection:} Creating multi-level mode hierarchies for fine-grained problem routing with sub-modes for specialized problem types
    \item \textbf{Cross-Modal Reasoning:} Extending CoS to vision-language tasks requiring coordination between visual and textual reasoning
    \item \textbf{Mode Composition:} Investigating how multiple modes can be combined within a single problem for complex tasks requiring diverse reasoning strategies
    \item \textbf{Theoretical Foundations:} Developing formal frameworks for understanding and predicting mode effectiveness based on problem characteristics
    \item \textbf{Adaptive Selection:} Learning to predict optimal k values for self-consistency based on problem difficulty and model confidence
\end{itemize}

\section{Conclusion}

We presented Chain of Simulation, a novel dual-mode reasoning framework that dynamically routes problems to specialized reasoning strategies in Large Language Models. By identifying and exploiting three distinct reasoning modes, computational flow for mathematics, symbolic state for spatial reasoning, and hybrid for multi-hop inference, CoS achieves superior performance across diverse benchmarks while maintaining computational efficiency. The comprehensive evaluation demonstrates that CoS achieves 71.5\% accuracy on GSM8K, 90.0\% on StrategyQA, and 19.0\% on bAbI, consistently outperforming strong baselines including Self-Consistency.

The framework offers better trade-offs between accuracy and efficiency, achieving comparable performance to Self-Consistency at less than half the computational cost. These results establish CoS as a practical approach for improving LLM reasoning in resource-constrained environments. The success of mode-specific reasoning reveals fundamental insights about LLM capabilities. Our analysis shows that correct mode selection is critical, computational mode achieves 81.2\% accuracy on appropriate problems but 0\% when misapplied. This finding suggests that effective reasoning in LLMs is as much about selecting the proper cognitive framework as it is about the reasoning process itself. Moreover, the existence of distinct, non-transferable reasoning modes implies that LLMs possess latent specialized capabilities that can be activated through targeted prompting. Chain of Simulation introduces a novel approach to improving LLM reasoning without additional training or architectural modifications. As models continue to scale, discovering and exploiting their latent reasoning modes will be crucial for achieving human-level performance across diverse cognitive tasks. 

This work demonstrates that careful analysis of problem characteristics combined with mode-specific prompting strategies can unlock hidden model capabilities, suggesting that the full potential of existing LLMs remains to be discovered. Looking forward, the principles underlying CoS, problem analysis, mode selection, and specialized execution may extend beyond reasoning to other aspects of LLM behavior. As we develop more sophisticated techniques for eliciting latent capabilities, we move closer to systems that can adapt their processing strategies to the task at hand, much as human cognition does. The journey toward artificial general intelligence may depend not just on scaling models but also on discovering how to effectively orchestrate their diverse latent capabilities.

\section*{Acknowledgment}
This work was supported by the Research Council of Finland through the 6G Flagship program (grant 318927), and the Emerging Projects program, Infotech Oulu.

\bibliographystyle{IEEEtran}
\bibliography{bib}

\end{document}